# Measuring the Impact of Scene Level Objects on Object Detection: Towards Quantitative Explanations of Detection Decisions

Lynn Vonder Haar, Timothy Elvira, Luke Newcomb, Omar Ochoa
Department of Electrical Engineering and Computer Science
Embry-Riddle Aeronautical University
Daytona Beach, FL
vonderhl@my.erau.edu, elvirat@my.erau.edu, newcombl@my.erau.edu, ochoao@erau.edu

**Abstract**

*Although accuracy and other common metrics can provide a useful window into the performance of an object detection model, they lack a deeper view of the model's decision process. Regardless of the quality of the training data and process, the features that an object detection model learns cannot be guaranteed. A model may learn a relationship between certain background context, i.e., scene level objects, and the presence of the labeled classes. Furthermore, standard performance verification and metrics would not identify this phenomenon. This paper presents a new black box explainability method for additional verification of object detection models by finding the impact of scene level objects on the identification of the objects within the image. By comparing the accuracies of a model on test data with and without certain scene level objects, the contributions of these objects to the model's performance becomes clearer. The experiment presented here will assess the impact of buildings and people in image context on the detection of emergency road vehicles by a fine-tuned YOLOv8 model. A large increase in accuracy in the presence of a scene level object will indicate the model's reliance on that object to make its detections. The results of this research lead to providing a quantitative explanation of the object detection model's decision process, enabling a deeper understanding of the model's performance.*

## 1. Introduction

The exact features that a Machine Learning (ML) model for object detection learns during training are not guaranteed. Predominant features will likely relate to the object in question, but secondary features may relate to the background of the object, or even its scene context. In many cases, a model learning scene level features has been found to be useful and improve object detection model performance and disambiguate the scene [1, 2, 3]. However, it is still important to understand the effects of these contextual features as they could pose limitations to the model, such as specific environments that increase or decrease the model's performance. Identifying the effect of those contextual features on the detection process can help to recognize these limitations and provide a better explanation for the detections made by the model.

The existing methods to identify the features that contribute to the detection process are known as explainability methods [4]. However, many of these methods involve the creation of heatmaps or other visuals that do not highlight exact, fine-grained features. Another helpful explanation for these detections would be quantitative, providing a more objective explanation for the detections made by the model. This paper presents a black box explainability method for object detection models based on the comparison of model accuracies when the testing dataset is distributed equally among various scene level objects. Additionally, this approach is able to pinpoint the positive or negative effects of scene level features the classifier learned to use to make a prediction. This research is not meant to reduce the model's reliance on context for object detection, but it aims to increase awareness of how context could be affecting the model's performance, either positively or negatively. Identifying the model's scene dependencies can give a unique reflection on the context nuances within the training data while also providing some explainability on scene-level contributions towards an inference. Providing explainability on the scene level objects allows the developer or user of the model to understand, to some extent, the model's performance of a classification when in the presence of a scene level object. Explainability is a critical non-functional quality attribute for all of ML; however, most methods are model-type specific, e.g., heatmaps. Explainability methods are also focused on the general inference and not inclusive of the smaller details which could have a large impact on confidence of the inference. This paper proposes a method designed to pinpoint subtle contextual details that can greatly enhance or deter the model's performance.

The organization of this paper is as follows. Section 2 presents background information on object detection, scene context in object detection, and explainability. Section 3 details the experiment that is the focus of this paper. Section 4 discusses the findings from the experiment in section 3. Section 5 reviews related work and outlines their key contributions and distinctly outlines how this paper's contributions differ. Section 6 outlines the future work. Section 7 concludes this paper and provides the authors' closing remarks.

## 2. Background

In many cases, context has been found to improve the performance of object detection models [1, 2, 3]. However, explanations of detection decisions are still necessary, especially in safety-critical domains. Context and explainability intersect in computer vision, namely object detection. Image level context refers to parts of an image that aggregate to understand the total understanding of the image. ML explainability can identify parts of the context within the image that are critical in its inference.

### 2.1. Object Detection

Object detection is a popular classification task for supervised learning. An object detection model trains on a class or set of classes, then identifies instances of those objects within an input image [5]. There are a few types of deep learning models that are popular for object detection tasks, including Convolutional Neural Networks (CNNs) and Region-based CNNs (R-CNNs) [6, 7]. CNNs are a type of artificial neural network often used in classification tasks due to their ability to filter the most important features in a more computationally efficient way than previous neural networks. The input to a CNN goes through several hidden layers, alternating between convolutional filters and pooling layers, to compute the pixel relevance to pre-defined classes [4]. Although initially used as image classifiers, CNNs can also be used for object detection by using a sliding window across the image and classifying any objects located within that sliding window [6]. However, a faster approach to object detection is with an R-CNN, which generates potential bounding boxes, then uses a classifier within those bounding boxes to refine their locations and dimensions [6, 7]. However, this requires the model to look at the image multiple times to generate and refine the bounding boxes, which still takes a significant amount of time. A more popular model now is the You Only Look Once (YOLO) model, which treats object detection as a regression problem, rather than a classification problem [6]. The model divides the image into a grid and any cell that includes the center of an object must detect that object with a confidence score, as is shown in Figure 1. This is done simultaneously for all objects in the image. In addition to the location, dimensions, and confidence predicted for each grid cell, there is a class-specific confidence score generated for each cell that describes how well the predicted bounding box matches the location and dimensions of the object, which can help refine the bounding box as needed. Throughout this process, the image is only run through the model once, making it much faster than the previous solutions.

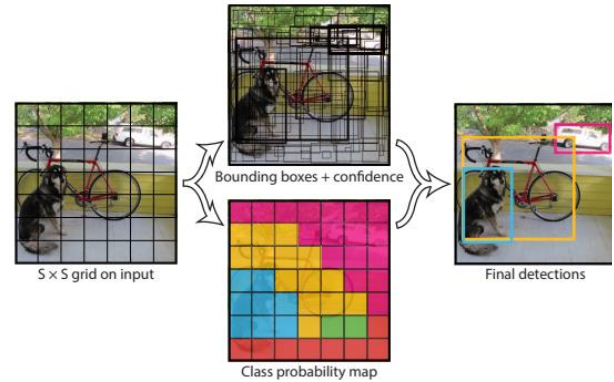

Figure 1. YOLO model. Adapted from [6].

Object detection models often struggle in the presence of occlusion or low image quality [8]. In these cases, models can learn to use the background context for an object to recognize its presence in the image.

### 2.2. Context in Object Detection

Neural networks are designed to imitate the processing occurring within the human brain [4, 9]. The human brain does not view objects in isolation but uses a rich understanding and recognition of scene context to identify objects, a method that has proven to be more efficient than identifying objects in isolation [8]. Therefore, some object detection models aim to imitate this ability [1, 6]. Additionally, context has been shown to improve performance with instances of poor image quality, noise, and major occlusions [1, 8].

Context, which is used synonymously with scene level objects in this paper, can be viewed in several ways including co-occurrence, relative location, and scale [1, 8]. Co-occurrence refers to objects that are semantically similar, or that often co-occur with each other. The presence of one of the objects indicates a higher chance that the other object will also be present in the image. Using the relative location of the related objects has also been shown to increase the efficiency of object detection [8]. By using relative location of objects, the model can focus on specific regions of the image. For instance, if a model is detecting vehicles, it is more likely to find the vehicle on a road than on the grass next to the road. Finally, some research suggests that scale, or size of objects relative to each other is a useful form of context as well [1]. For instance, a car that is shorter than a child may be more likely to be a toy than a vehicle on the road. Examples of how each of these context types can be used in object detection are shown in Figure 2.

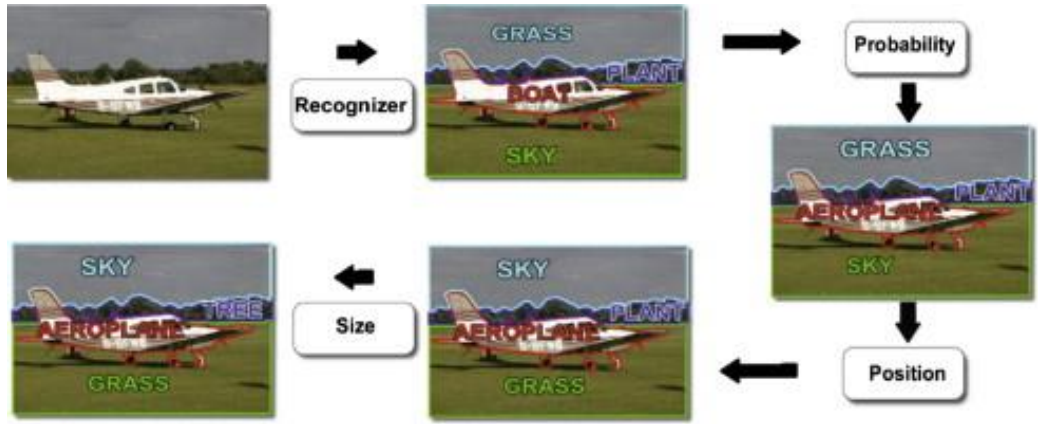

Figure 2. Context-based categorization. Adapted from [1].

Context has been used by ML developers to improve the performance of object detection models [6]. Some developers use context to overcome occlusion or low resolution in images while still maintaining high model performance [10, 11]. In some cases, using context can add beneficial side effects to the developing model, such as improvements to the model's robustness and performance [6, 12]. However, despite its popular use in object detection, there is still a need to identify context's impact on a model and scrutinize any limitations that it may produce in the model.

### 2.3. Explainability

Explainability in ML is the ability to explain its reasoning for predictions or inferences for a given input [13, 14]. Explainability is about making it easier to understand how ML models make their inferences; the domain is expansive and encompasses many types of ML, birthing an entire field dedicated to eXplainable Artificial Intelligence (XAI) [15]. This is not to be confused with interpretable AI which are simplistic models that have an output inherently understandable by humans [16]. There are two main types of explainability methods: black-box and white-box approaches [17].

Black-box explainability approaches aim to explain the decision making of the ML models which have complex or obfuscated innerworkings that are not easily understood by humans [18]. There is an entire taxonomy of black-box methods which digress into model-specific and model agnostic methods. Due to variations and nuanced complexities of certain models, there are specific explainability methods designed solely for that model paradigm. For example, Pixel-wise Relevance Propagation (PRP), is a model-specific explainer that is used to apply relevance scores to pixels in images for a convolutional neural network. A more model agnostic approach would be Local Interpretable Model-agnostic Explanations (LIME) uses random samples from the neighborhood of the inference to generate feature importance vectors [18]. Because LIME is model-agnostic, it can be used to explain any black-box model compared to PRP which requires a specific model to use. Black-box explainability methods produce meaningful explanations; however, because the explanations are not directly associated with the model architecture, there is risk of the explanations being disconnected with the true rationale behind the model's inference.

Because of black-box explainability lacking transparency and connectivity to model internal components, developers employ white-box methods to generate explanations directly involved with the model's mechanisms [19]. White-box explainability methods are predominantly applied to image classification tasks. Some examples of these explainability tasks are deconvolutional networks, guided/gradient backpropagation, and SmoothGrad [19]. Additionally, there are many visual analytics and explanations for the white-box methods, namely PRP which adds a relevance score for nodes across the network, giving the user a visual heatmap for a given input as it progresses through forward propagation [20]. White-box methods illustrate the process or identify areas that contribute to a given model's inference. Both black-boxes and white-boxes have their advantages and disadvantages; white-box methods open the model to scrutinize its decision but are not as versatile or widely applicable as black-box methods.

Current explainability methods have proven helpful in understanding the decisions that ML models make. However, many explainability methods for image tasks, such as object detection, output a heatmap of which areas of the image were most helpful in the decision-making process and which areas were not helpful. An example of this is shown in Figure 3, which is an example of a heatmap generated by PRP [21]. As can be seen in Figure 3, heatmaps generated by current explainability methods can be open to interpretability by the person viewing them. This makes the precision of using these heatmaps difficult.

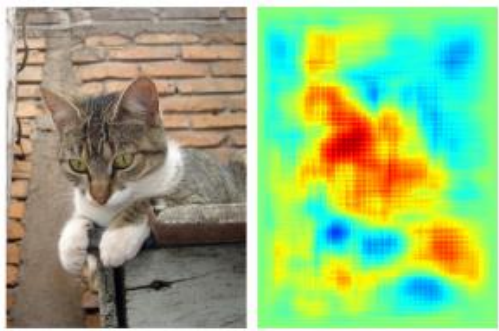

Figure 3. Heatmap generated through PRP. Adapted from [21].

## 3. Experiment

The purpose of this research is to propose a new black-box explainability method that provides a quantitative and global explanation for object detection. Not only does this proposed explainability method reduce the subjectivity of the results interpretation, but it also provides explanations regarding the global context of the image, rather than individual pixels and their neighbors. This is accomplished by training a model, then testing it on four different testing datasets that have potential scene level object contributions and comparing the model's accuracies. The steps of the experiment are as follows and are depicted in Figure 4:

1. Train the model on a set of classes.
2. Distribute the testing dataset to match the class distribution of the training dataset.
3. Test the model.
4. Compile testing datasets with the same class distribution, but that include certain scene level objects that are heavily present in the training dataset.
5. Test the model.
6. Compare the model accuracies to determine the effect of the scene level objects on the model's performance.

The images used in this experiment are of emergency road vehicles, specifically emergency medical services (EMS) vehicles, fire vehicles, police vehicles, mobile communication vehicles, rescue vehicles, and tow trucks. The application of this experiment could be in explaining model detections of emergency vehicles in autonomous driving situations. The images were collected through Google Images by searching for "emergency road vehicles", or the individual vehicle desired. A total of 394 images were collected and split into training and validation datasets. The training data was annotated using Roboflow's object detection mode [22]. The data was annotated into the six classes mentioned previously.

Transfer learning was applied to the YOLOv8 model using this training dataset. The dataset was exported directly from Roboflow to fine-tune the YOLOv8 model. The training was completed over 80 epochs with a patience of 50, meaning that the model will stop training if it does not detect a significant performance improvement for 50 consecutive epochs. The model completed all 80 training epochs.

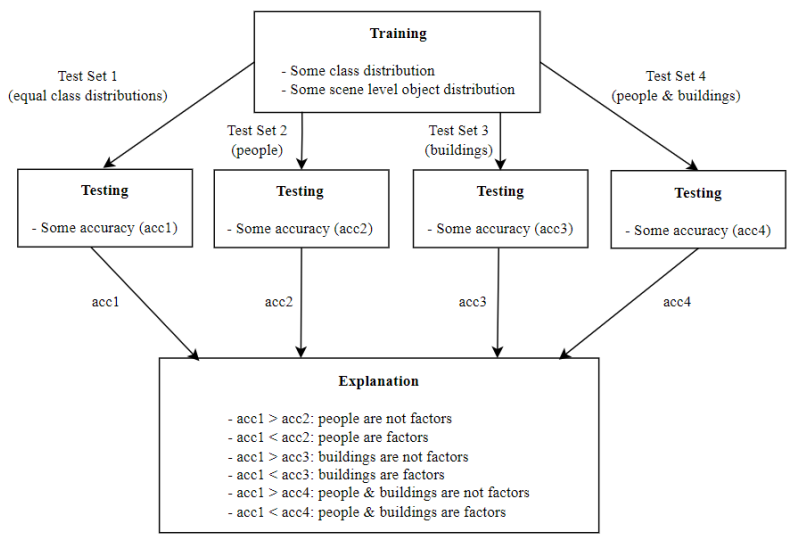

Figure 4. The experimental process.

The first step of the experiment was to run the fine-tuned YOLOv8 model on a normal, randomized test set that matched the class distribution of the training dataset. Therefore, the distribution of EMS vehicles, fire vehicles, mobile communication vehicles, police vehicles, rescue vehicles, and tow trucks in the testing dataset matched the distribution of those same classes within the model's training dataset. This testing dataset did not focus on any scene level object. It was generated automatically by Roboflow when the annotated data was split into training, validation, and testing datasets. The testing dataset consisted of 43 images with a class distribution matching that of the training dataset. This distribution was manually confirmed. After running the testing data through the model, the accuracy was found to be 63.00%. It is expected that more training data would increase this accuracy. However, the goal of this experiment was to compare the accuracies between the four testing datasets, so the experiment focused on the relative accuracies and was therefore unaffected by the absolute accuracies of the model.

The second part of this experiment involved creating new test datasets for each of the scene level objects analyzed, as well as one where both scene level objects appeared in the images. The purpose of this part of the experiment was to analyze the effect of the chosen scene level objects on the performance of the object detection model. In this experiment, buildings and people were chosen as the scene level objects in question because they were heavily present in the training dataset and therefore were hypothesized to have a high probability of affecting the detection decisions of the fine-tuned YOLOv8 model. For this part of the experiment, three separate test datasets were compiled as follows:

1. A class distribution matching the training dataset with each image containing at least one person.
2. A class distribution matching the training dataset with each image containing at least one building.
3. A class distribution matching the training dataset with each image containing at least one person and at least one building.

Each of the test datasets consisted of 43 images with the same class distribution as the training dataset and the first, randomized testing dataset. Once each of the three test datasets were compiled, they were run through the same fine-tuned YOLOv8 model as the first testing dataset.

|  | Normal test set | People | Buildings | People & Buildings |
|---|---|---|---|---|
| Number of correct detections | 34 | 25 | 40 | 38 |
| Number of vehicles to detect | 54 | 45 | 48 | 58 |
| Accuracy | 63.00% | 55.56% | 83.33% | 65.52% |
| Contributes to detection? | N/A | No | Yes | Yes |

Table 1. Experimental results.

The comparison of the test results is shown in Table 1. As seen in Table 1, the accuracy of the model on each test set was calculated based on the number of vehicles that should have been detected and number of vehicles that were correctly detected. This process was done manually. As shown in Table 1, there was a decrease in accuracy for the testing dataset that included people, indicating that the presence of a person in the image did not improve the performance of the model and was potentially even negatively impacting the ability of the model to detect the emergency vehicles. There was little increase in accuracy for the testing dataset that included both people and buildings, suggesting that the presence of both scene level objects in each image may be positively contributing to the performance of the model, but not by much. However, there was a significant increase in accuracy for the testing dataset that included buildings, specifically an increase in accuracy from 63.00% to 83.33%, suggesting that buildings were largely contributing to the model making correct detection decisions. In addition to the increase in accuracy, many of the detections made in the buildings test set had higher confidence scores than the normal test set. There was also significantly less detection confusion when buildings were present. There were fewer mistakes on the dimensions on the bounding boxes and there were fewer instances of double detections or misdetections. These observations support the results of the experiment, suggesting that buildings largely contributed to the model's correct detections of emergency road vehicles.

## 4. Discussion

The experiment presented in this paper showed great success in identifying the impact of specified scene level objects on an object detection model's performance. The array of results for the different scene level objects demonstrates how this method can recognize both the positive and negative contributions of the context. This can help developers and users of an object detection model understand their model's performance and certain limitations of their model based on context.

Not only did some of the scene level objects appear to contribute to the accuracy of the model's detections, but they also seemed to positively impact the confidence scores with which the model made its detections. The presence of buildings in one of the testing datasets seemed to help the model detect emergency road vehicles correctly and with higher confidence than the randomized testing dataset. Additionally, there was substantially less detection confusion in the testing dataset with buildings present. This means that there were fewer errors in the dimensions of the bounding boxes, as well as fewer overlapping detections for the same object.

Although the experiment had many successes, there are some limitations to the experiment presented in this paper. When analyzing the results of the experiment, it is important to discuss the nondeterminism of ML models. Although these were the results for this specific fine-tuned YOLOv8 model, the same results may not be achieved if the experiment is repeated. For instance, if the YOLOv8 model was fine-tuned again with the same training data, the accuracies may be different. Additionally, if measures are taken to reduce the contribution of buildings to this model's detection process through further fine-tuning of the model, the experiment will need to be repeated to determine if the new model still reflects the same context reliance.

Another limitation is that the accuracies achieved during this experiment were low, but it is expected that more training data would increase those accuracies. However, since this experiment was focused solely on the relative accuracies of the model on a normal testing dataset versus testing datasets with scene level objects, the absolute accuracies of the model were not factors.

Regarding calculating the accuracy of the model on each of the testing datasets, there was detection confusion that was difficult to take into account at times. There were times when the class was correct, but there was a discrepancy in the dimensions of the bounding box. There were also times when two separate classes were detected for the same object. These limitations were difficult to account for in a single accuracy score but given that the same approach was taken for each of the testing datasets, the relative accuracies are still meaningful.

Despite these limitations, the results of the experiment are still promising. Since the experiment analyzed the relative accuracies of the model between each of the testing datasets and the model's accuracy on each of the testing datasets was calculated in the same way, the results are meaningful even with the limitations of the study.

The research presented in this paper was meant as a new black-box explainability method to identify reliance on scene level objects. However, this paper does not analyze the effectiveness or usefulness of object detection models utilizing context to make detections. This paper also does not present solutions to model context reliance. The explainability method in this experiment can help developers and users of object detection models understand the limitations of their model's performance, specifically constraints of certain scene level objects increasing or decreasing the model's performance.

## 5. Related Work

The role of context in machine learning computer vision is a long-standing area of research. Context has been categorized into several groups: interposition, support, probability, position, and size [1, 23]. Previous research has detailed and provided different types of contexts to object detection models in order to better understand its role in accurately detecting objects. This was mostly focused on extracting non-semantic context such as detecting the weather or geographical location of the image [24]. Some have used scene-level objects to detect other objects which would otherwise be difficult to identify due to their size or blending into their surroundings such as a baseball bat in the hands of a baseball player or a computer mouse on a desk [25]. These studies exemplify that context is an important consideration in object detection, however fewer studies have been done in attempting to explain the precise relationship between context and model accuracy.

In autonomous vehicle vision systems, the focus is on explaining how context influences predictions for safety reasons [26, 27, 28]. Some authors have used an algorithm that outputted the parts of the image that were stressed by the object detection model as a heat map, which they used as an explainability approach to show the model's attention [26, 28]. Others had a similar focus on autonomous vehicles and compared the impact of the class distribution and model attention on the accuracy of a model on multiple training datasets [27]. This was once again done using a heatmap explainability approach.

Other papers have worked to identify the effects of context on object detection in domains outside of autonomous driving [29, 30]. As in autonomous driving, some authors suggest using heatmaps, such as class activation maps [30]. However, another paper analyzed the impact of context on object detection by testing the model's accuracy on variations of the image [29]. Some variations included masking the object, blurring the background, and masking the background. By analyzing the model's ability or inability to correctly classify the object with these variations, the authors were able to note the impact of context on the model.

A similar paper presents a methodology for removing associated scene-level objects from training images in order to both measure the effect on an existing model, but also to improve the robustness of the model when detecting objects alone [31]. This method applied augmentation of the dataset to both object detection and image segmentation and aimed to quantify the relationships before and after. It showed that by removing context from some of the training data they were able to reduce reliance on context without losing performance, and the model was still able to learn context. The researchers demonstrated that the original baseline model was over reliant on context and would make false predictions in situations that could be safety-critical. The analysis of the reliance on context in a quantitative way is similar to the goal of this paper, showing specific interest in improving explainability in this way.

Another paper proposed a method to introduce interpretability into object detection models [32]. Specifically, the authors applied their interpretability method to Faster R-CNN models, which have a different architecture from the YOLOv8 model used in this paper. They did this through the latent structures of the model to understand the regions of interest, or the attention of the model. This paper's focus was slightly different than the papers on explainability because it focused on an end-to-end understanding of the model. Interpretability attempts to provide transparency of the model's inner workings, whereas explainability simply seeks to understand an individual decision. However, the authors still sought to better understand the performance of the object detection model, as was done in this paper.

## 6. Future Work

The experiment presented in this paper provided promising results for a new black-box explainability method for object detection models. However, the accuracies achieved in the experiment were low. It is expected that additional training samples would improve the model's accuracy. While this did not affect the experiment since this research was focused on the accuracies relative to each other, some future work could be dedicated to repeating this experiment with more training samples to increase the model's overall accuracy.

The empirical findings of this experiment are still rudimentary. Future steps for this work include experimentation with other mathematical or statistical expressions that can express the impact in a formalized approach. This preliminary work aims at comparing the accuracy of two images, one with the classification and the other with the classification and context object, to find the observational impact of other objects within a scene.

Additionally, the experiment presented in this paper analyzed the impact of two scene level objects separately and together on the model's performance. Future research could include repeating this experiment with other scene level objects to ensure the translation of this method to many different contexts. This experiment could be repeated with the same dataset, but looking at different scene level objects, or even in an entirely new domain.

Another experiment could also repeat the steps described here with different object detection model architectures, such as CNNs and R-CNNs. This direction of future work could aim to verify differences in the usage of context in different model architectures. YOLO was specifically designed to utilize context within the image to improve performance of the model while only looking at the image once [6]. Therefore, it would be interesting to compare the results of this experiment with those of other object detection models.

The purpose of this research was to identify if certain scene level objects were contributing to the detections made by a fine-tuned YOLOv8 model. However, this paper does not analyze the usefulness of an object detection model utilizing context to make its detections. Although it is beneficial to know how context affects a model's detections, the developer may choose to maintain this behavior due to real life co-occurrence. Future research could be done into the usefulness and effectiveness of maintaining this behavior in certain applications.

This paper also does not include solutions for context reliance. Another potential area of future work is reducing the effects of context on the detection process. Removing images with buildings from the training data or adding additional training instances without buildings present may help to reduce this reliance. Other solutions may include occlusion of certain scene level objects, or other methods of forcing the attention of the model to focus on the object itself rather than its context.

## 7. Conclusion

The purpose of this paper was to present a new black box explainability method for object detection models. The method presented provides a quantitative explanation by comparing the model's accuracy on multiple testing datasets with and without certain scene level objects. Not only is this a quantitative explanation, but it also provides a global explanation for the detection process, rather than a pixel-level view. The experiment performed in this paper analyzed the effect of people and buildings on the detection process of emergency road vehicles by a fine-tuned YOLOv8 model. The experiment included testing the fine-tuned model on four different datasets: a standard randomized dataset, a dataset where all images included at least one person, a dataset where all images included at least one building, and a dataset where all images included at least one person and at least one building. By comparing the model's accuracy on each of these testing datasets, it was found that people, and people and buildings did not have an effect on the performance of the model. However, buildings did have a large effect on the performance of the model, suggesting that the context one or more buildings improved the model's ability to correctly detect emergency road vehicles.

The results of this research can help developers and users of object detection models understand the limitations of their model's performance regarding the presence of certain scene level objects in their data. Potential areas of future research to this study include repeating the experiment with more data and other new scene level objects, analyzing the effectiveness of context reliance in certain applications, and finding solutions to context reliance for applications where it hinders the effectiveness of the detection process. This experiment could also be repeated with other object detection model architectures to analyze the translatability of the explainability method to other architectures.